\documentclass[10pt,twocolumn,letterpaper]{APSIPA2021}

\usepackage{times}
\usepackage{epsfig}
\usepackage{graphicx}
\usepackage{amsmath}
\usepackage{amssymb}
\usepackage{multicol}
\usepackage{subfigure}
\usepackage{cite}

\usepackage[pagebackref=true,breaklinks=true,letterpaper=true,colorlinks,bookmarks=false]{hyperref}
\usepackage{caption}

\usepackage{geometry}
\usepackage{fancyhdr}

\fancypagestyle{firststyle}{
  \fancyhf{}
  \fancyhead[C]{2023 Asia Pacific Signal and Information Processing Association Annual Summit and Conference (APSIPA ASC)}
}

\begin{document}

\title{Marine Snow Removal Benchmarking Dataset}
\author{
\authorblockN{
Reina Kaneko\authorrefmark{1},
Yuya Sato\authorrefmark{2},
Takumi Ueda\authorrefmark{2},
Hiroshi Higashi\authorrefmark{1}, and
Yuichi Tanaka\authorrefmark{1}\thanks{This work is supported in part by Japan Science and Technology Agency (JST) Advanced International Collaborative Research Program (AdCORP) under Grant JPMJKB2307 and JSPS KAKENHI under Grant 23K17461.}
}
\\
\authorblockA{
\authorrefmark{1}
Osaka University, Japan \\
\authorrefmark{2}
 Tokyo University of Agriculture and Technology, Japan \\}

\authorblockA{
E-mail: r.kaneko@sip.comm.eng.osaka-u.ac.jp, \{higashi, ytanaka\}@comm.eng.osaka-u.ac.jp}

}

\maketitle
\thispagestyle{firststyle}
\pagestyle{fancy}

\begin{abstract}
This paper introduces a new benchmarking dataset for marine snow removal of underwater images.
Marine snow is one of the main degradation sources of underwater images that are caused by small particles, e.g., organic matter and sand, between the underwater scene and photosensors.
We mathematically model two typical types of marine snow from the observations of real underwater images.
The modeled artifacts are synthesized with underwater images to construct large-scale pairs of ground truth and degraded images to calculate objective qualities for marine snow removal and to train a deep neural network.
We propose two marine snow removal tasks using the dataset and show the first benchmarking results of marine snow removal.
The \emph{Marine Snow Removal Benchmarking Dataset} is publicly available online.
\end{abstract}

\section{Introduction}
\label{sec:intro}
Image restoration has been one of the main topics of computer vision for decades \cite{zhangGaussianDenoiserResidual2017a,chan2016plug,romano2017little,xie2012image,chen2018deep,xu2014deep,yeh2017semantic}.
In the current era of deep learning, the quality of image restoration for various tasks has been significantly improved under a sufficient number of pairs of ground truth and degraded images.
Accordingly, the current focus on image restoration will be for images taken under extreme situations such as underwater, satellite, and medical images \cite{akkaynak2018revised, berman2017diving, li2019underwater, ueda2019underwater,wangDeepCNNMethod2017,li2020underwater, dudhane2020deep,jiang2020novel,chen2014vehicle,zhang2016deep,zhu2017deep,maggiori2016convolutional,shen2017deep,litjens2017survey,weigert2018content,anwarDivingDeeperUnderwater2020}.
This problem is called \emph{extreme image restoration} herein.
Extreme images often have an untypical degradation beyond that of a Gaussian model, and the numbers of available images for training and modifying restoration algorithms are also limited.
As a result, extreme image restoration is a challenging problem.

In this paper, we focus on underwater image enhancement.
Its main challenge is mostly due to the fact that we generally have no ground truth images.
This makes it difficult to evaluate the objective qualities of the restored images and measure the restoration performance.
Moreover, the lack of pairs of ground truth and degraded image causes a difficulty in training a neural network (if we consider using a deep neural network for restoration) because the loss functions of the image restoration typically evaluate some objective image quality metrics such as mean squared error (MSE) and structured similarity index (SSIM) \cite{zhouwangImageQualityAssessment2004}.
Hence, unsupervised restoration methods, including linear and nonlinear filtering, are still popular in underwater image enhancement \cite{berman2017diving, li2019underwater,akkaynak2019sea}.

Several studies for underwater image enhancement with deep neural networks have been proposed \cite{ueda2019underwater,wangDeepCNNMethod2017,li2020underwater, dudhane2020deep,jiang2020novel,9775132,9426457,9854113}.
Most focus on removing the color shift from underwater images.
That is, the blueish underwater images are restored by enhancing the red and green channels.
Conventional underwater image enhancement methods using deep neural networks tackle the problem on the lack of large-scale underwater image datasets in various ways.
For example, the studies in \cite{wangDeepCNNMethod2017,anwarDeepUnderwaterImage2018,ueda2019underwater,dudhane2020deep} synthetically generate  degraded images from clean images taken on the ground by simulating the degradation processes of absorption and scattering.
Moreover, \cite{jiang2020novel,liWaterGANUnsupervisedGenerative2018,fabbriEnhancingUnderwaterImagery2018} use generative adversarial networks.
And a number of challenges to improving the restoration performance still exist, some promising results have been presented.
However, such studies ignore another main source of degradation for underwater images: \textit{Marine snow}.

\begin{figure}[t]
\begin{center}
   \subfigure{\includegraphics[width = .4 \linewidth]{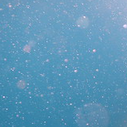}}
   \subfigure{\includegraphics[width = .4 \linewidth]{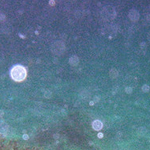}}
\end{center}
   \caption{Marine snow artifacts in real underwater images.}
\label{realMarineSnow}
\end{figure}

In this paper, we consider a marine snow removal (MSR) problem for underwater images by constructing a large-scale dataset containing underwater image pairs.
Marine snow is a typical underwater image degradation, examples of which are shown in Fig. \ref{realMarineSnow}.
Marine snow artifacts are visible in digital images when we take an underwater image using a flashlight.
Small particles in the scene, e.g., organic matter and sand, cause marine snow.
The particles reflect the light from the flashlight, and the reflected light will be captured by the photosensors.
Marine snow is often annoying in underwater photography because it affects the overall image quality.
However, the particles are unevenly distributed in the scene, and the degradation process is difficult to model completely.
Therefore, standard methods for marine snow removal are still limited to simple median-filter-based approaches \cite{banerjeeEliminationMarineSnow2014,cyganekRealtimeMarineSnow2018,koziarskiMarineSnowRemoval2019, 9775132}.
Moreover, as with standard underwater image enhancement, we do not have a pair of ground truth and degraded images to calculate the objective image qualities and tune the parameters of the restoration algorithms.

We propose marine snow synthesis methods for solving the above-mentioned problems of MSR.
In other words, we \textit{intentionally} append the marine snow artifacts to real underwater images without marine snow and generate pairs of ground truth images and degraded images (i.e., clean underwater images and their corresponding images with marine snow artifacts).
The key for synthesizing artifacts is to mathematically model marine snow in digital images as accurately as possible.
From actual marine snow images, we discover that many marine snow artifacts can be modeled through two typical pixel value distributions.
By utilizing this finding, we can synthetically generate many pairs of underwater images with/without marine snow artifacts.

We also propose two marine snow removal tasks and develop datasets corresponding to these tasks.
This \textit{Marine Snow Removal Benchmarking Dataset} (MSRB Dataset) is the first attempt in this field and is publicly available online\footnote{\url{https://github.com/ychtanaka/marine-snow}}.
We also present the first MSR benchmarking results using the MSRB Dataset: We compare the objective performance of marine snow removal among median filters and a deep neural network.
Real MSR results as well as the limitations are also described.

The remainder of this paper is organized as follows:
Section \ref{sec:relatedworks} reviews several related studies for underwater image enhancement.
Two representative marine snow models that we discovered are presented in Section \ref{sec:marine_snow_models} along with their synthesizing method.
The MSRB Dataset specifications are introduced in Section \ref{sec:dataset} and include the description of the two tasks.
The first MSR benchmarking results with objective image quality metrics are shown in Section \ref{sec:benchmark}.
Real MSR results and limitations are described in Section \ref{sec:real_msr}.
Finally, we provide some concluding remarks in Section \ref{sec:conclusion}.

\section{Related Studies}
\label{sec:relatedworks}
There have been few studies on modeling marine snow in digital images.
Its seminal study is described in \cite{boffety2012phenomenological, boffety2012color}.
A marine snow artifact is modeled using a Gaussian function in which the artifact is less transparent in its center than in its surrounding area.
This method is better than the traditional salt and pepper noise; however, it does not simulate the actual shapes of marine snow artifacts.
As a result, the center part of marine snow becomes excessively thick and the generated images are far from the real ones.

MSR is also an underrepresented problem in underwater image enhancement.
Here, we briefly review some existing approaches.

A widely used method for MSR is a median filter (MF) \cite{brownrigg1984weighted, huang1979fast}.
If marine snow artifacts can be assumed sufficiently small (typically $1$--$3$ pixels in diameter), MF is expected to work because small marine snow artifacts can be considered to have a salt-and-pepper effect.
However, it significantly blurs the entire image, particularly when we use a large filter kernel size.

Specific to MSR, a modified version of a MF is proposed in \cite{banerjeeEliminationMarineSnow2014,farhadifardSingleImageMarine2017}.
This method applies the MF selectively if the target pixel has a higher intensity than the surrounding pixels.
However, it is still difficult to remove large marine snow artifacts.

A few MSR methods for video sequences have also been proposed.
They utilize the fact that marine snow artifacts continuously move in consecutive video frames.
In \cite{farhadifardMarineSnowDetection2017}, background modeling is used to remove marine snow artifacts from a static scene.
Marine snow artifacts are tracked and a customized MF is applied to the detected artifacts in \cite{cyganekRealtimeMarineSnow2018}.
However, they are not applicable to a single underwater image.
Importantly, all methods above are based on a model-based approach.
This is mainly due to a lack of high-quality datasets, as mentioned in Section \ref{sec:intro}.

A deep learning-based MSR method is proposed in \cite{9775132}.
Its training dataset consists of synthesized marine snow images, corresponding ground truth images, marine snow masks. This dataset also contains real underwater images with marine snow (without ground truth).
Marine snow particles are manually made with Adobe Photoshop: That results in a lack of randomness and scalability.

In the following, we address the contributions of this study; marine snow models and the dataset specifications.

\begin{figure}[t]
    \centering
    \subfigure[][Enlarged portion]{\includegraphics[width = .4 \linewidth]{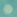}}
    \subfigure[][3D plot]{\includegraphics[width = .4 \linewidth]{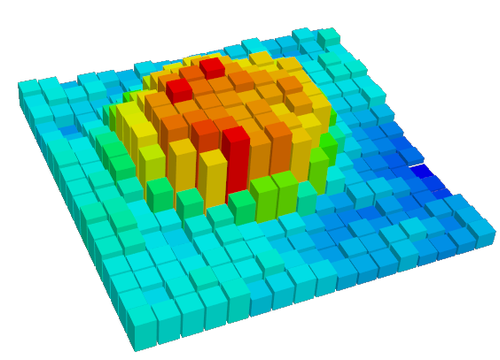}}\\
    \subfigure[][Enlarged portion]{\includegraphics[width = .4 \linewidth]{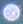}}
    \subfigure[][3D plot]{\includegraphics[width = .4 \linewidth]{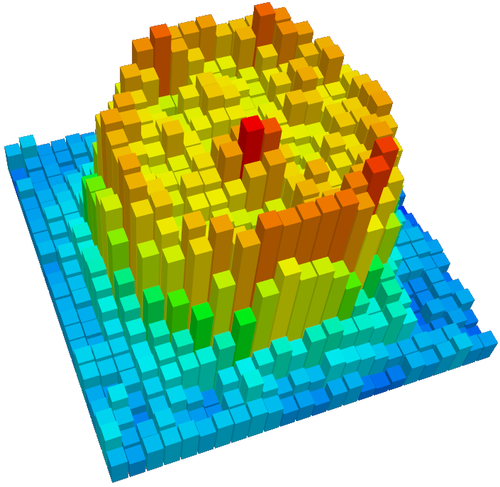}}
    \caption{Marine snow examples in real underwater images. Top: Highland type (type H). Bottom: Volcanic Crater type (type V). The 3D plots correspond to the grayscale images.}
    \label{crater_real_img}
\end{figure}

\section{Marine Snow Models}
\label{sec:marine_snow_models}
There are various sources of marine snow and it is impractical to estimate the sources of all  particles from a single underwater image.
Instead of the estimation of the sources, we model the pixel value distributions of marine snow artifacts from observations of underwater images.

\subsection{Real Marine Snow Examples}
First, we start by showing marine snow examples in real underwater images.
Figs. \ref{crater_real_img}(a) and (c) show enlarged portions of representative marine snow artifacts cropped from real underwater images.
Although they look similar, their pixel value distributions are slightly different.

Taking a closer look, the 3D plots of Figs. \ref{crater_real_img}(a) and (c) are shown in Figs. \ref{crater_real_img}(b) and (d), respectively.
As clearly observed, they do \textit{not} have a shape like a Gaussian function in contrast to the conventional assumption in \cite{boffety2012color,boffety2012phenomenological}.
Rather than a Gaussian function, these 3D plots are similar to \textit{elliptic conical frusta}, i.e., sliced elliptic cones.
Furthermore, the top surfaces of the frusta have different characteristics between Figs. \ref{crater_real_img}(b) and (d).
In our preliminary observation, most marine snow artifacts can be classified into these two representative shapes.
In the following, we present the mathematical models of marine snow artifacts that reflect the above-mentioned observations.

\subsection{Marine Snow Model 1: Highland Type}
The first marine snow model is the \textit{Highland Type} (type H).
The type H marine snow corresponds to Fig. \ref{crater_real_img}(a) and can be modeled as an elliptic conical frustum with a rough surface.

\begin{figure}[t]
\centering
   \includegraphics[width=0.7\linewidth]{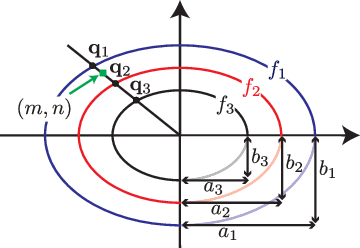}
   \caption{Ellipses used for marine snow synthesis.}
\label{three_ellipses}
\end{figure}

For modeling type H marine snow, we first suppose to have two ellipses $f_1$ and $f_2$, both centered at the coordinates $(k, l)$.
As shown in Fig. \ref{three_ellipses}, we assume the focal points of $f_1$ and $f_2$ are located on the horizontal axis, i.e., the ellipses are wider, for simplicity.
Let $a_i$ and $b_i$ be the semi-major and semi-minor axes of $f_i$ where $i\in \{1,2\}$, respectively.
Suppose that $f_1$ is larger than $f_2$, i.e., $a_1>a_2$ and $b_1 > b_2$ and the eccentricities of $f_1$ and $f_2$ are identical, i.e., $\frac{a_1}{b_1} = \frac{a_2}{b_2}$.

For notational simplicity, $(k,l)$ is set to $(0,0)$ hereafter.
The $(m,n)$th pixel value of the type H marine snow $h(m,n)$ is formulated as follows:
\begin{equation}
    h(m,n) =
    \begin{cases}
    c + \epsilon & \frac{m^2}{a_2^2} + \frac{n^2}{b_2^2} \le 1\\
    c\frac{d(\mathbf{p}, \mathbf{q}_1)}{d(\mathbf{q}_1, \mathbf{q}_2)} + \epsilon & 1 < \frac{m^2}{a_2^2} + \frac{n^2}{b_2^2} \land \frac{m^2}{a_1^2} + \frac{n^2}{b_1^2} \le 1\\
    0 & \text{otherwise},
    \end{cases}
    \label{eqn:type_h}
\end{equation}
where $c$ is a constant that determines the transparency of the marine snow, $d(\cdot, \cdot)$ calculates the Euclidean distance between two points, and $\epsilon$ is a small perturbation that mimics a rough surface.
Furthermore, $\mathbf{p}:=[m,n]^\top$ and $\mathbf{q}_i$
is the intersection of the ellipse $f_i$ and the straight line crossing $(m,n)$, as illustrated in Fig. \ref{three_ellipses}.

\begin{figure}[t]
    \centering
    \subfigure[][Modeled artifact]{\includegraphics[width = .4 \linewidth]{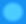}}
    \subfigure[][3D plot]{\includegraphics[width = .4 \linewidth]{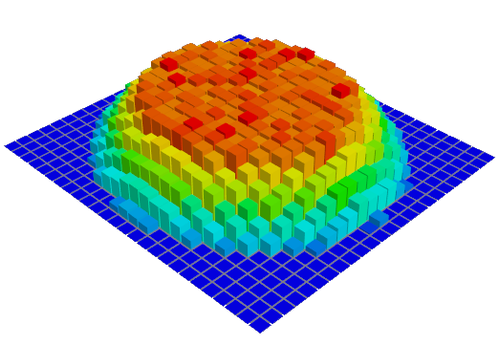}}\\
    \subfigure[][Modeled artifact]{\includegraphics[width = .4 \linewidth]{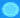}}
    \subfigure[][3D plot]{\includegraphics[width = .4 \linewidth]{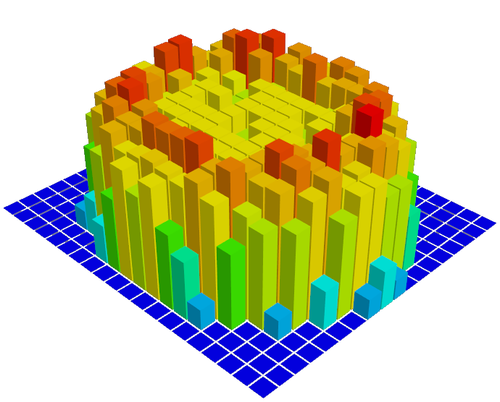}}
    \caption{Proposed marine snow models. Top: Type H marine snow. Bottom: Type V marine snow.}
    \label{artificial_highland}
\end{figure}

An example of the synthesized type H marine snow artifacts is shown in Fig. \ref{artificial_highland}(a) along with its 3D plot in Fig. \ref{artificial_highland}(b).
See the similarity in Figs. \ref{crater_real_img}(a) and (b).
For creating a dataset, the major axis of $h(m,n)$ is randomly rotated and the transparency $c$ is also randomly chosen.
These specifications are shown later in Section \ref{sec:dataset}.

\begin{table*}[t]
    \caption{Parameter settings for marine snow synthesis where $\mathcal{U}(v_{\min}, v_{\max})$ is a continuous uniform distribution between $v_{\min}$ and $v_{\max}$.}
    \label{tab:params}
    \centering
    \begin{tabular}{c|c}
    \hline
    Parameter & Setting\\\hline
    $c$ in \eqref{eqn:type_h}     &  $\mathcal{U}\{10, 80\}^\ast$\\
    $c_r$ in \eqref{eqn:type_v} & $\mathcal{U}\{1, 40\}^\ast$\\
    Gaussian filter size in \eqref{eqn:image_syn}     & $(2r+1) \times (2r+1)$ where $r \sim \mathcal{U}(1,6)$\\
    $a_2$ & $\begin{cases}
    \mathcal{U}(1, 3) & \text{for small marine snow artifacts}\\
    \mathcal{U}(3, 16) & \text{for large marine snow artifacts}
    \end{cases}$\\
    $b_2$ & $\begin{cases}
    \mathcal{U}(0.3a_2, a_2) & \text{for small marine snow artifacts}\\
    \mathcal{U}(0.8a_2, a_2) & \text{for large marine snow artifacts}
    \end{cases}$\\
    $a_1$ & $\mathcal{U}(a_2, 2a_2)^\dagger$\\
    $a_3$ & $\mathcal{U}(0.5a_2, a_2)^\ddagger$\\\hline
    \multicolumn{2}{l}{$\dagger$ $b_1$ is obtained from the eccentricity of $f_2(m,n)$}\\
    \multicolumn{2}{l}{$\ddagger$ $b_3$ is obtained from the eccentricity of $f_2(m,n)$}
    \end{tabular}
\end{table*}
\subsection{Marine Snow Model 2: Volcanic Crater Type}
The second marine snow model is the volcanic crater type (type V).
An example of type V marine snow is shown in Figs. \ref{crater_real_img}(c) and (d).
This can be modeled by a modified version of the type H with an overshoot top edge of the frustum.

For the type V marine snow, we consider another ellipse $f_3$, where $a_3$ and $b_3$ are its semi-major and semi-minor axes, respectively.
It is located inside $f_2$, as illustrated in Fig. \ref{three_ellipses}.
As for the type H, focal points of all ellipses are assumed to be located along the horizontal axis
and their eccentricities are identical, i.e., $\frac{a_1}{b_1} = \frac{a_2}{b_2} = \frac{a_3}{b_3}$.

Based on the type H function $h(m,n)$, the $(m,n)$th pixel value of the type V marine snow $v(m,n)$ is formulated as $v(m,n) = h(m,n) + g(m,n)$,
where $g(m,n)$ is the \textit{rim} term defined as
\begin{equation}
\begin{split}
    &g(m,n)\\
& =     \begin{cases}
    c_r (1\! -\! \frac{d(\mathbf{p}, \frac{1}{2}(\mathbf{q}_2 + \mathbf{q}_3))}{\frac{1}{2} d(\mathbf{q}_2, \mathbf{q}_3)}) &\!\!  1 < \frac{m^2}{a_3^2} + \frac{n^2}{b_3^2} \land \frac{m^2}{a_2^2} + \frac{n^2}{b_2^2} \le 1\\
    0 & \text{otherwise,}\\
    \end{cases}
\end{split}
\label{eqn:type_v}
\end{equation}
where $c_r$ is the constant for the maximum rim height and $\mathbf{q}_3$ indicates the coordinates of the intersection of the ellipse $f_3$ and the straight line crossing $(m,n)$.

The synthesized type V marine snow artifact is shown in Figs. \ref{artificial_highland}(c) and (d), which are similar to the real versions in Figs. \ref{crater_real_img}(c) and (d).

\subsection{Synthesizing Images with Marine Snow}
By using the marine snow models introduced in the previous subsections, we synthesize the marine snow with underwater images.

Let $U \in \mathbb{R}^{H\times W \times 3}$ be an original underwater image.
First, we randomly select a target pixel coordinate $(k, l)$ to append the marine snow.
Second, we randomly select the major and minor axes of the ellipses $a_i$ and $b_i$.
In addition, the direction of the major axis is randomly set.
Third, for each channel, type H or V marine snow is independently generated, where
the parameters $c$ and $c_r$ are independently set.
We denote an image with one marine snow artifact as $M$, where pixel values in $M$ are all-zero except the marine snow artifact.
Finally, $U$ and $M$ are combined as follows to yield the synthesized image $I$:
\begin{equation}
    I_{k,l,t} = 
    \begin{cases}
    M_{k,l,t} + G_{k,l,t} & M_{k,l,t} \ge 0\\
    U_{k,l,t} & \text{otherwise}
    \end{cases}
    \label{eqn:image_syn}
\end{equation}
where $G \in \mathbb{R}^{H\times W \times 3}$ is the Gaussian filtered version of $U$ that reflects blur owing to marine snow.

We iterate the process until a predetermined number of marine snow artifacts are appended.

\section{Dataset Specifications}
\label{sec:dataset}
In this section, we present the specifications of synthesized marine snow artifacts in the MSRB Dataset.
It is designed for two tasks:
\begin{description}
    \item[Task 1]: Removal of small-sized marine snow artifacts.
    \item[Task 2]: Removal of various-sized marine snow artifacts.
\end{description}
Clearly, Task 2 is more difficult than Task 1.

First, detailed parameters and setups shared in both tasks are presented.
Second, we introduce MSR Tasks with corresponding synthesized images.

\begin{figure*}[tp]
\centering
    \subfigure{\includegraphics[width = 0.132\linewidth]{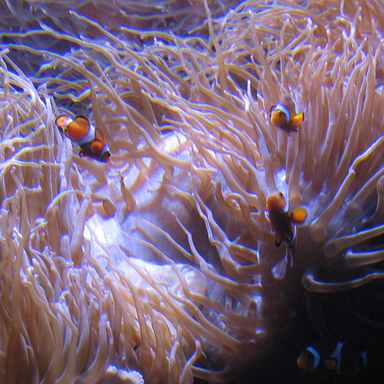}} 
    \subfigure{\includegraphics[width = 0.132\linewidth]{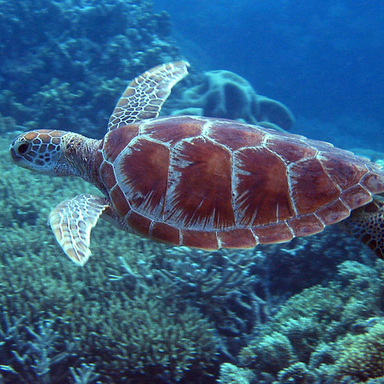}} 
    \subfigure{\includegraphics[width = 0.132\linewidth]{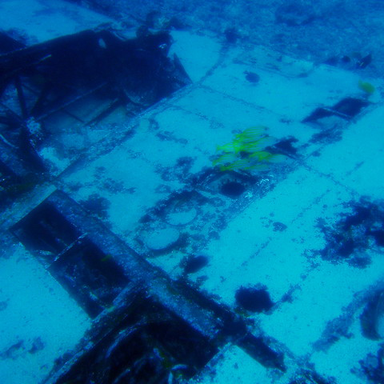}}
    \subfigure{\includegraphics[width = 0.132\linewidth]{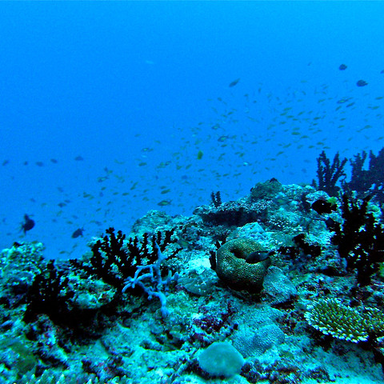}} 
    \subfigure{\includegraphics[width = 0.132\linewidth]{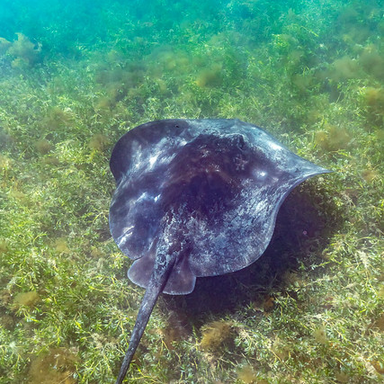}}\\
    \vspace{-0.15in}\caption*{Original underwater images.}
    \subfigure{\includegraphics[width = 0.132\linewidth]{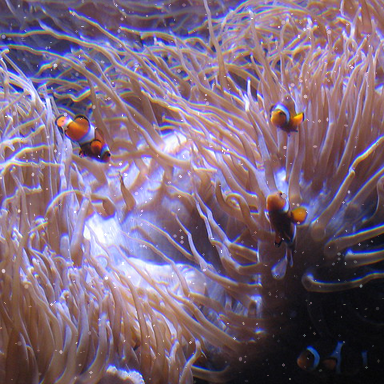}} 
    \subfigure{\includegraphics[width = 0.132\linewidth]{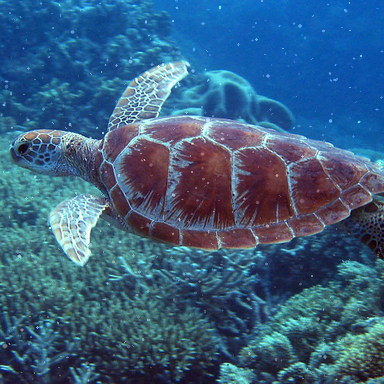}} 
    \subfigure{\includegraphics[width = 0.132\linewidth]{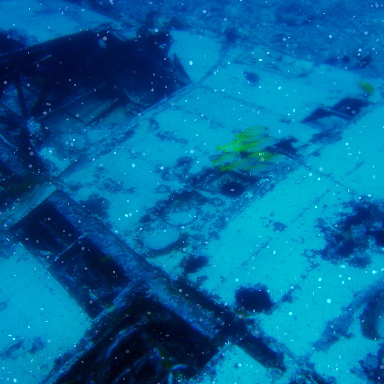}}
    \subfigure{\includegraphics[width = 0.132\linewidth]{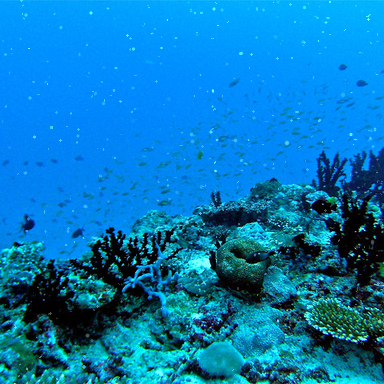}} 
    \subfigure{\includegraphics[width = 0.132\linewidth]{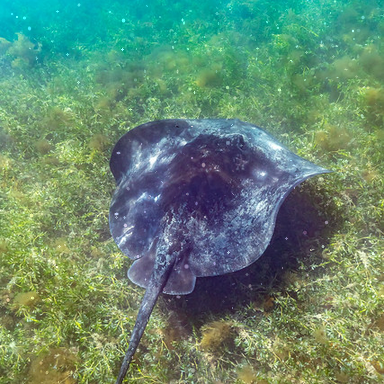}} \\
    \vspace{-0.15in}\caption*{Synthesized images for MSR Task 1.}
    \subfigure{\includegraphics[width = 0.132\linewidth]{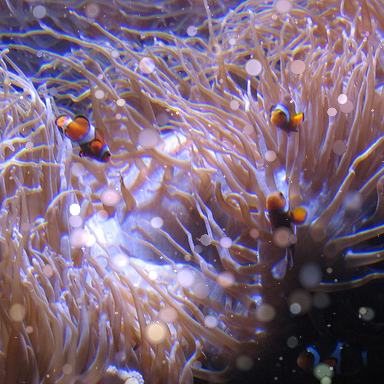}} 
    \subfigure{\includegraphics[width = 0.132\linewidth]{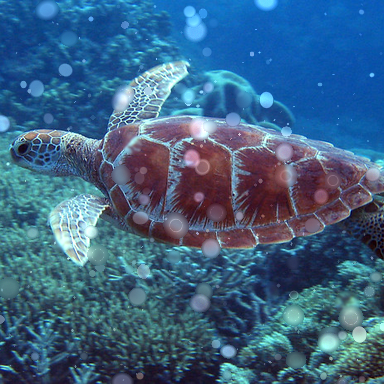}} 
    \subfigure{\includegraphics[width = 0.132\linewidth]{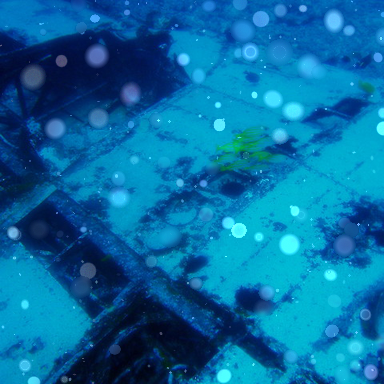}} 
    \subfigure{\includegraphics[width = 0.132\linewidth]{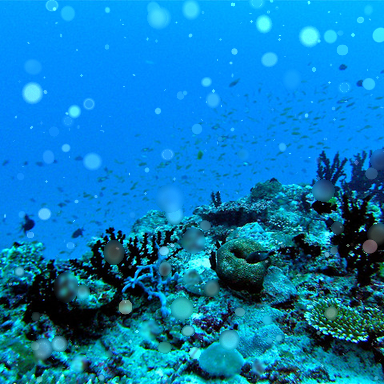}} 
    \subfigure{\includegraphics[width = 0.132\linewidth]{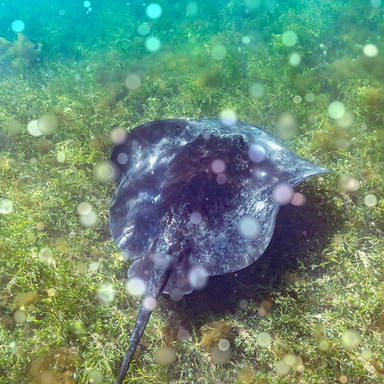}}\\
    \vspace{-0.15in}\caption*{Synthesized images for MSR Task 2.}
  \caption{Example images in MSRB dataset.}
  \label{dataset_img}
\end{figure*}

\begin{figure*}[t]
\centering
    \subfigure{\includegraphics[width = 0.132\linewidth]{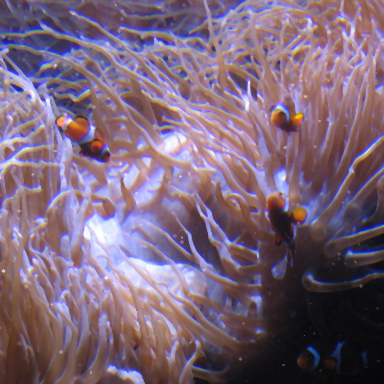}} 
    \subfigure{\includegraphics[width = 0.132\linewidth]{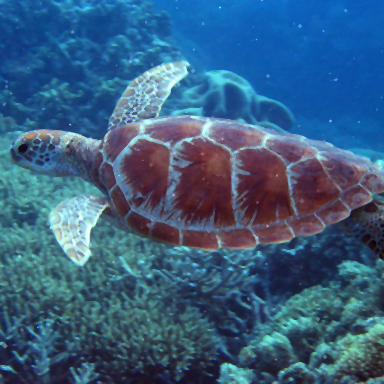}} 
    \subfigure{\includegraphics[width = 0.132\linewidth]{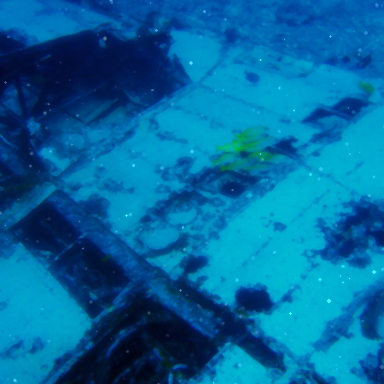}}
    \subfigure{\includegraphics[width = 0.132\linewidth]{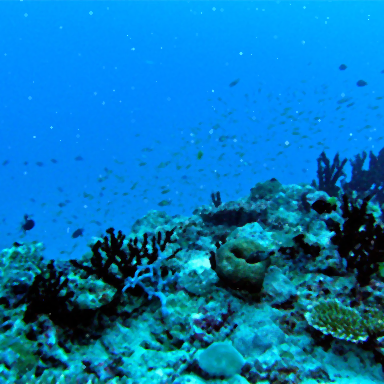}}
    \subfigure{\includegraphics[width = 0.132\linewidth]{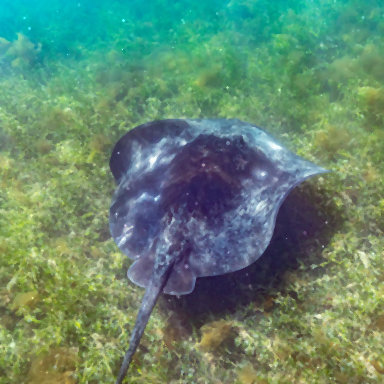}} \\
    \vspace{-0.15in}\caption*{Restoration results by MF.}
    \subfigure{\includegraphics[width = 0.132\linewidth]{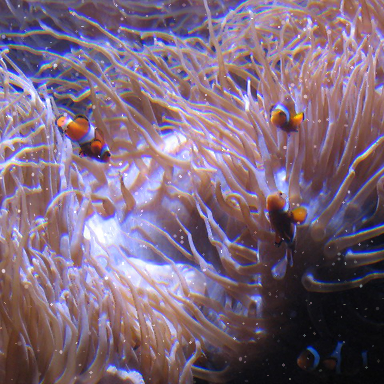}} 
    \subfigure{\includegraphics[width = 0.132\linewidth]{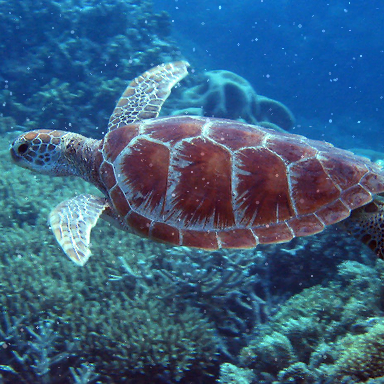}} 
    \subfigure{\includegraphics[width = 0.132\linewidth]{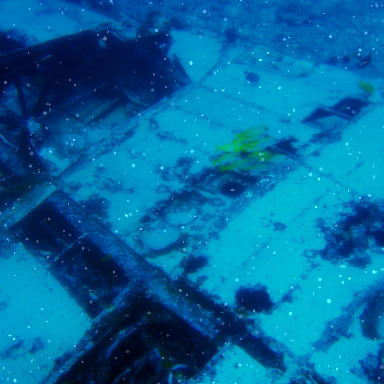}}
    \subfigure{\includegraphics[width = 0.132\linewidth]{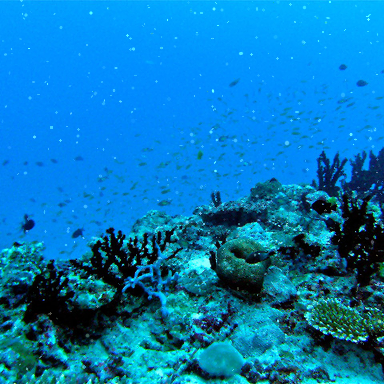}} 
    \subfigure{\includegraphics[width = 0.132\linewidth]{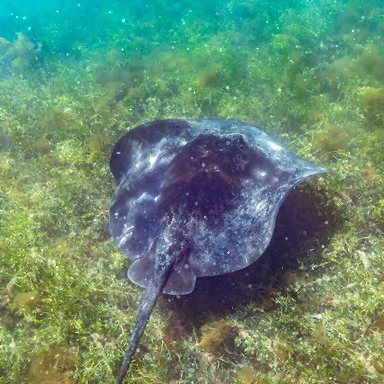}} \\
    \vspace{-0.15in}\caption*{Restoration results by adaptive MF.}
    \subfigure{\includegraphics[width = 0.132\linewidth]{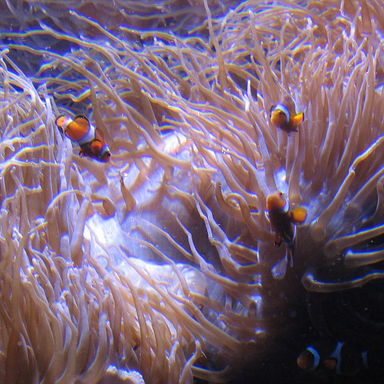}} 
    \subfigure{\includegraphics[width = 0.132\linewidth]{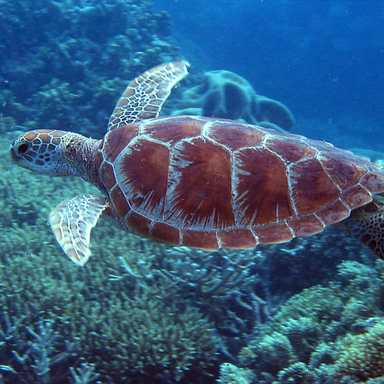}} 
    \subfigure{\includegraphics[width = 0.132\linewidth]{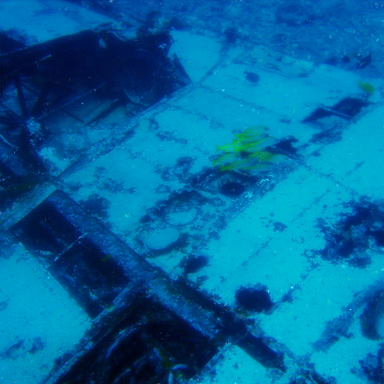}} 
    \subfigure{\includegraphics[width = 0.132\linewidth]{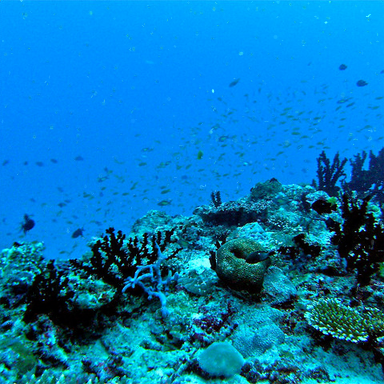}} 
    \subfigure{\includegraphics[width = 0.132\linewidth]{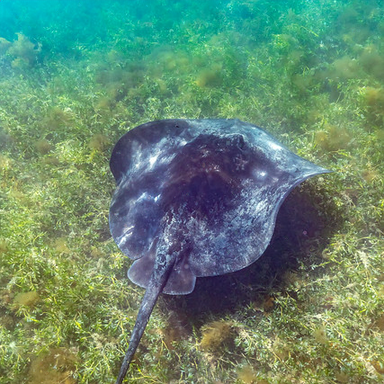}}\\
    \vspace{-0.15in}\caption*{Restoration results by U-Net.}
  \caption{MSR Task 1 results.}
  \label{small_validation}
\end{figure*}

\begin{figure*}[t]
\centering 
    \subfigure{\includegraphics[width = 0.132\linewidth]{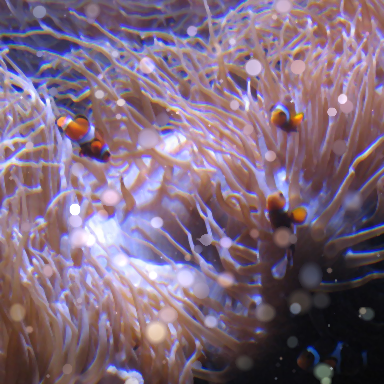}}
    \subfigure{\includegraphics[width = 0.132\linewidth]{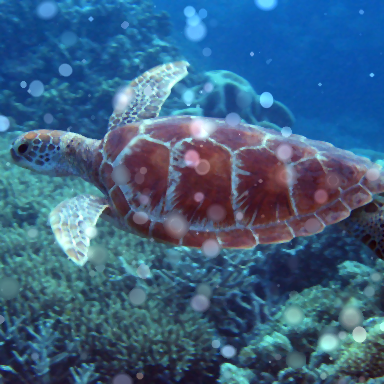}}
    \subfigure{\includegraphics[width = 0.132\linewidth]{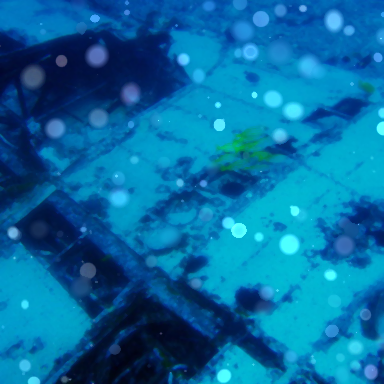}} 
    \subfigure{\includegraphics[width = 0.132\linewidth]{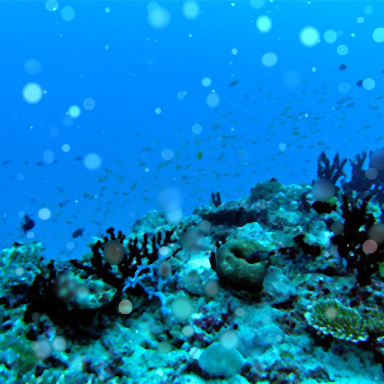}} 
    \subfigure{\includegraphics[width = 0.132\linewidth]{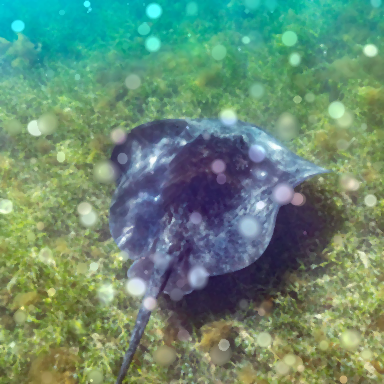}} \\
    \vspace{-0.15in}\caption*{Restoration results by MF.}
    \subfigure{\includegraphics[width = 0.132\linewidth]{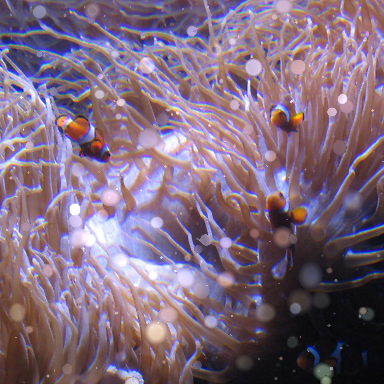}} 
    \subfigure{\includegraphics[width = 0.132\linewidth]{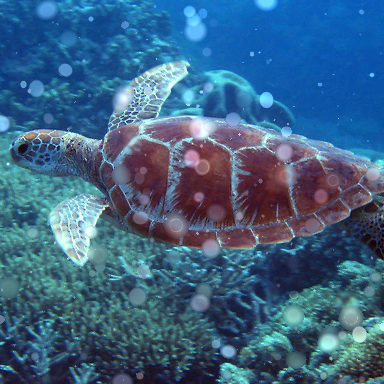}} 
    \subfigure{\includegraphics[width = 0.132\linewidth]{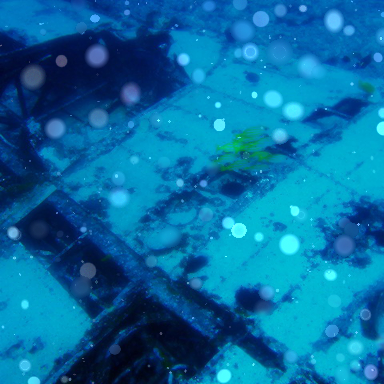}} 
    \subfigure{\includegraphics[width = 0.132\linewidth]{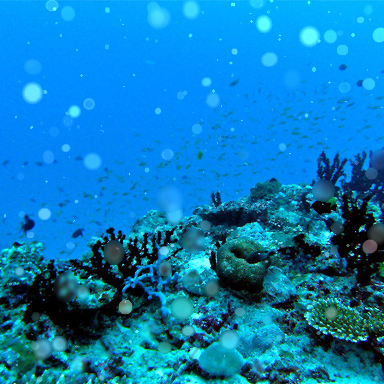}} 
    \subfigure{\includegraphics[width = 0.132\linewidth]{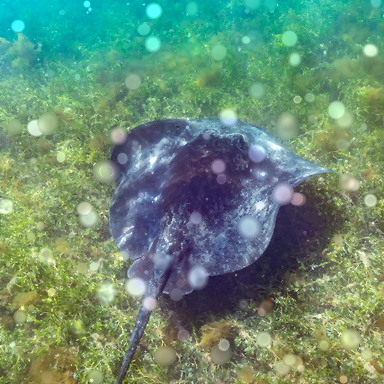}} \\
    \vspace{-0.15in}\caption*{Restoration results by adaptive MF.}
    \subfigure{\includegraphics[width = 0.132\linewidth]{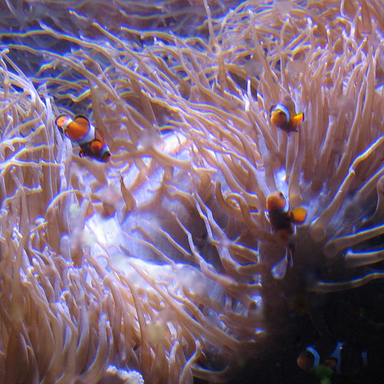}} 
    \subfigure{\includegraphics[width = 0.132\linewidth]{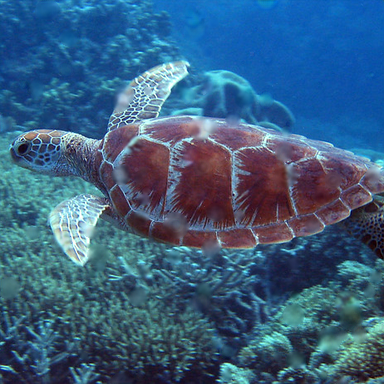}} 
    \subfigure{\includegraphics[width = 0.132\linewidth]{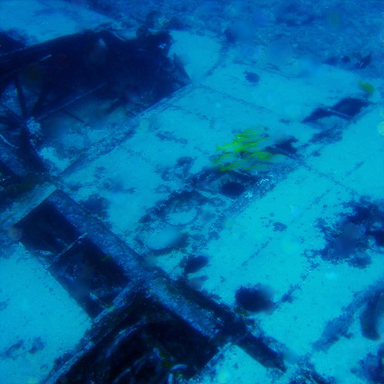}} 
    \subfigure{\includegraphics[width = 0.132\linewidth]{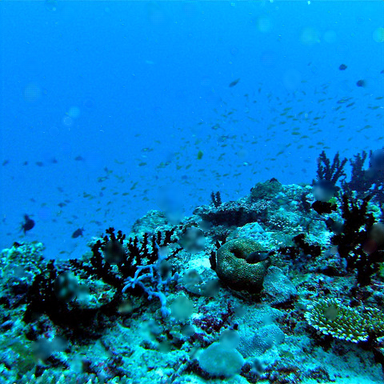}} 
    \subfigure{\includegraphics[width = 0.132\linewidth]{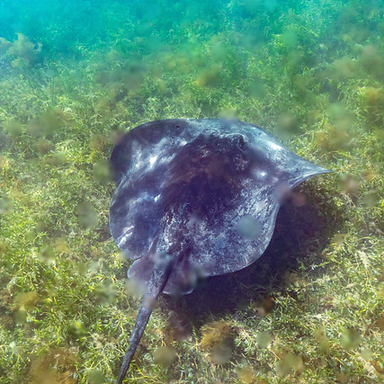}}\\
    \vspace{-0.15in}\caption*{Restoration results by U-Net.}
  \caption{MSR Task 2 results.}
  \label{combined_validation}
\end{figure*}

\subsection{General Setup}
Each sub-dataset corresponding to an MSR Task contains $2,300$ training image pairs and $400$ test image pairs, all having a pixel resolution of $384 \times 384$.
An image pair contains one original underwater image and one image containing synthesized marine snow artifacts.
All original images are collected from flickr\footnote{\url{https://www.flickr.com/}} under a Creative Commons Attribution-NonCommercial-ShareAlike 2.0 Generic (CC BY-NC-SA 2.0) License.

Each synthesized image contains $N$ marine snow particles where $N$ is chosen from $\mathcal{U}\{100, 600\}$, in which $\mathcal{U}\{v_{\min}, v_{\max}\}$ is a discrete uniform distribution between $v_{\min}$ and $v_{\max}$.
In each synthesized image, type H and V marine snow particles are randomly generated with a probability of $0.7$ for type H and $0.3$ for type V, according to our preliminary observations.
The other representative parameters introduced in the previous section are listed in Table \ref{tab:params}.
Most of the parameters are determined according to our preliminary observations of real underwater images with marine snow artifacts.

\subsection{MSR Task 1: Removing Small-Sized Marine Snow Artifacts}
We introduce the first MSR Task: Removal of small-sized marine snow artifacts.
In Task 1, the maximum width/height of the artifacts is restricted to $6$ pixels, which corresponds to roughly $1.6$\% of the image width/height.
Task 1 is designed for underwater scenes where particles are relatively far from the photosensors.
Because the task is relatively simple, the conventional MF approach is expected to be successfully applied (at least partially).

Examples of test image pairs in MSR Task 1 are shown in Fig. \ref{dataset_img}.
As clearly visualized, the synthesized images have various sized (but small) marine snow artifacts.

\subsection{MSR Task 2: Removing Various-Sized Marine Snow Artifacts}
The second MSR task is designed for underwater scenes containing particles in various distances.
As a result, the synthesized images for Task 2 have small- and large-sized marine snow artifacts.
For large-sized artifacts, we set the largest width/height of marine snow to $32$ pixels, which corresponds to $8.3$\% compared to the image width.
Furthermore, the probabilities of small- and large-sized artifacts are set to $0.7$ and $0.3$, respectively.

Training image examples for Task 2 are also shown in Fig. \ref{dataset_img}.
The images have various-sized artifacts and the degraded areas are larger than the images in Task 1.

In the following, we show the benchmarking results of MSR as well as the real MSR.

\section{MSR Benchmarking Results}
\label{sec:benchmark}
In this section, we present the first benchmarking performance for MSR Tasks 1 and 2.
The methods used for the benchmarking are 1) MF \cite{huang1979fast}, 2) adaptive MF \cite{banerjeeEliminationMarineSnow2014}, and 3) U-Net \cite{ronnebergerUNetConvolutionalNetworks2015}.
The kernel size of MFs is set to $3\times 3$ or $5 \times 5$ pixels.
Thanks to the MSRB Dataset, we can train a deep neural network for removing (synthesized) marine snow.
For a benchmarking purpose, we train U-Net \cite{ronnebergerUNetConvolutionalNetworks2015} using the dataset.
The parameter settings are slightly different from the original.

The MSR results for Tasks 1 and 2 are shown in Figs. \ref{small_validation} and \ref{combined_validation}, respectively.
As clearly observed, MF oversmoothes the images, which results in blur even for regions without marine snow artifacts.
The adaptive MF suppresses blur; however, many artifacts are not removed mainly owing to the threshold-based algorithm.
Furthermore, both MFs do not operate successfully for large marine snow artifacts in Task 2.
By contrast, U-Net successfully removes marine snow for both tasks.
Even for Task 2, it is able to suppress the large marine snow artifacts because of its multiresolution structure and the large number of trainable parameters.

For an objective comparison, we compute the average PSNRs and SSIMs over the test datasets.
We summarize the results in Table \ref{task_1_psnr_ssim}.
The objective measure indicates that U-Net is superior to the MFs for both MSR tasks.
Note that all MFs present smaller PSNRs/SSIMs than those of the synthesized images because they filtered areas without marine snow.
By contrast, U-Net achieves a better objective quality than the synthesized images, which implies the effect of a deep neural network for MSR.

\begin{table}[t]
  \centering
  \caption{Average PSNRs and SSIMs over the test dataset.}
  \label{task_1_psnr_ssim}
  \begin{tabular}{l|rr||rr} \hline
  & \multicolumn{2}{c||}{Task 1} & \multicolumn{2}{c}{Task 2}\\\hline
  Method & PSNR & SSIM & PSNR & SSIM\\ \hline
  MF (3 $\times$ 3) & 28.55 & 0.846 & 22.81 & 0.770\\
  MF (5 $\times$ 5) & 25.98 & 0.711& 21.93 & 0.645\\
  Adaptive MF (3$\times$3) & 29.88 & 0.910& 23.35 & 0.842\\
  Adaptive MF (5$\times$5) & 28.08 & 0.861& 22.83 & 0.794\\
  U-Net & \bf{36.82} & \bf{0.978} & \bf{30.95} & \bf{0.932}\\\hline
    Synthesized image& 32.20 & 0.945 & 23.83 & 0.876\\\hline
  \end{tabular}
\end{table}

\section{Real Marine Snow Removal Examples}
\label{sec:real_msr}
We construct the MSRB Dataset to mimic real marine snow artifacts.
In this section, we show the real MSR results and present limitations of the dataset.

\subsection{Real Marine Snow Removal}
\begin{figure}[tp]
\vspace{-20pt}
\centering
    \subfigure{\includegraphics[width = 0.26\linewidth]{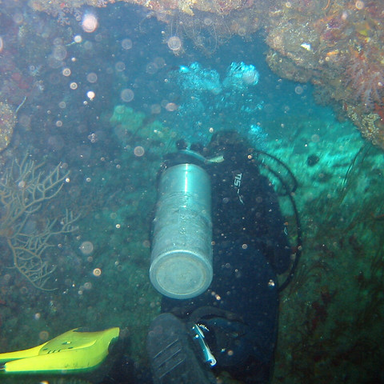}}
    \subfigure{\includegraphics[width = 0.26\linewidth]{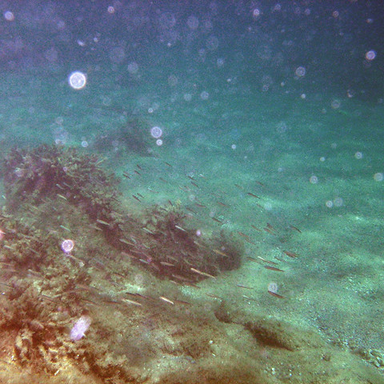}}
    \subfigure{\includegraphics[width = 0.26\linewidth]{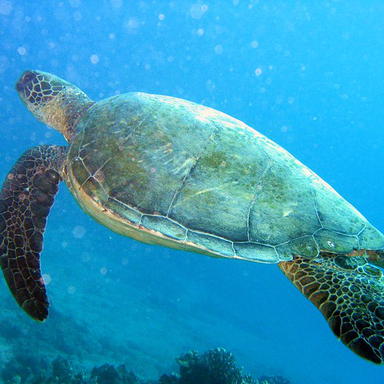}}\\
    \vspace{-0.15in}\caption*{Real underwater images.}\vspace{-0.05in}
    \subfigure{\includegraphics[width = 0.26\linewidth]{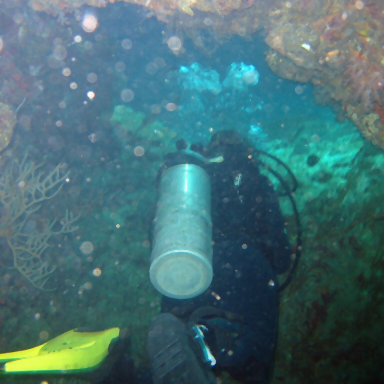}}
    \subfigure{\includegraphics[width = 0.26\linewidth]{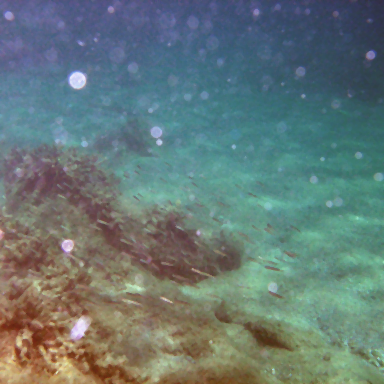}}
    \subfigure{\includegraphics[width = 0.26\linewidth]{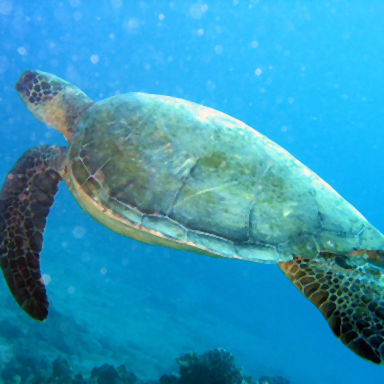}}\\
    \vspace{-0.15in}\caption*{Restoration results by MF.}\vspace{-0.05in}
    \subfigure{\includegraphics[width = 0.26\linewidth]{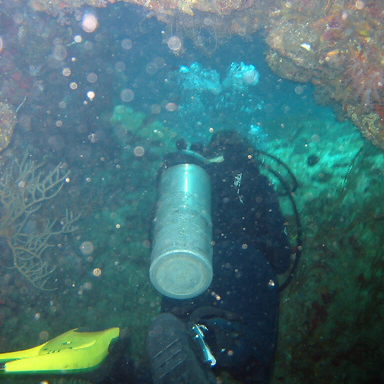}}
    \subfigure{\includegraphics[width = 0.26\linewidth]{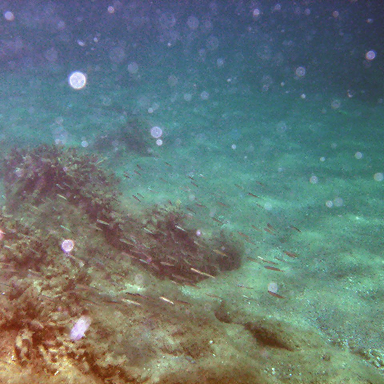}}
    \subfigure{\includegraphics[width = 0.26\linewidth]{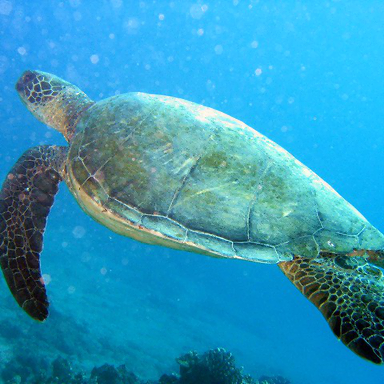}}\\
    \vspace{-0.15in}\caption*{Restoration results by adaptive MF.}\vspace{-0.05in}
    \subfigure{\includegraphics[width = 0.26\linewidth]{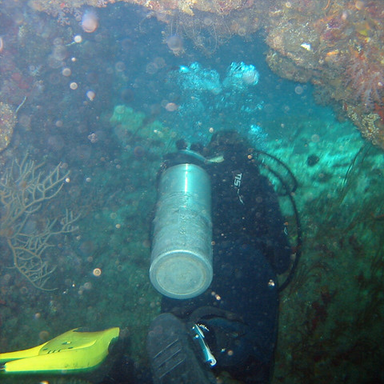}}
    \subfigure{\includegraphics[width = 0.26\linewidth]{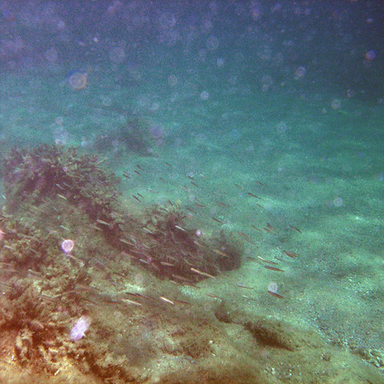}}
    \subfigure{\includegraphics[width = 0.26\linewidth]{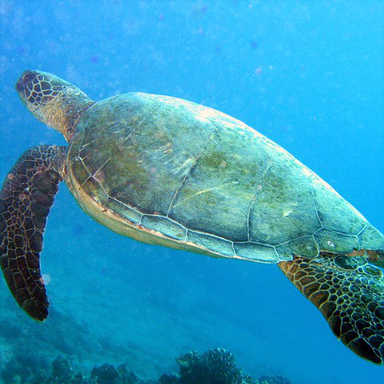}}\\
    \vspace{-0.15in}\caption*{Restoration results by U-Net.}\vspace{-0.05in}
    \subfigure{\includegraphics[width = 0.26\linewidth]{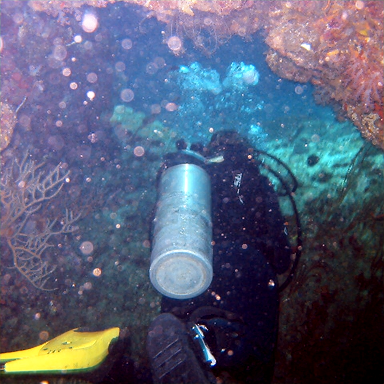}}
    \subfigure{\includegraphics[width = 0.26\linewidth]{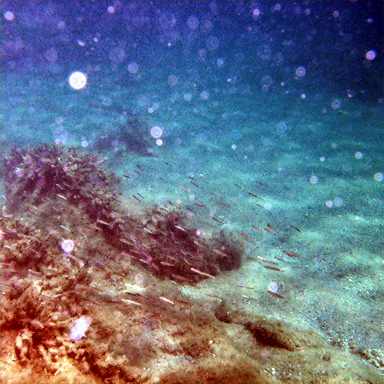}}
    \subfigure{\includegraphics[width = 0.26\linewidth]{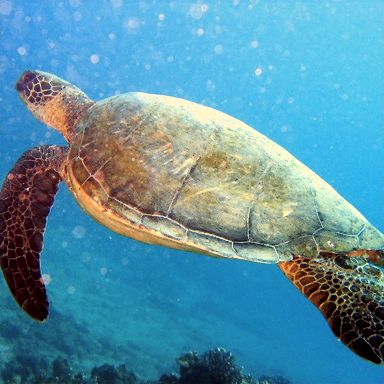}}\\
    \vspace{-0.15in}\caption*{Restoration results by Deep WaveNet.}\vspace{-0.05in}
  \caption{Real MSR results.}\vspace{-0.2in}
  \vspace{-20pt}
  \label{real_img_elimination}
\end{figure}

Fig. \ref{real_img_elimination} shows MSR results for some real images.
We use MF, adaptive MF, and U-Net as in the previous section. In addition, Deep WaveNet \cite{sharma2023wavelength} is applied to validate the regular underwater enhancement model works for marine snow removal.
U-Net is trained using MSRB Task 2 Dataset.
The trained model of Deep WaveNet is taken from the original authors' GitHub repository.

As clearly observed, MFs cannot suppress marine snow artifacts because of their various sizes and different transparencies. Deep WaveNet also fails to remove marine snow artifacts. That means marine snow cannot be suppressed by underwater image enhancement techniques alone.
U-Net presents a better restoration image quality than MFs: It (partly) suppresses artifacts having both small and large sizes.
This implies the effectiveness of deep-learning-based methods trained using the MSRB Dataset for real MSR.

Note that thick marine snow artifacts remain in the restored images.
The reason for this could be two-fold: First, U-Net is not specifically designed for MSR although we slightly customized its structure.
A neural network designed specifically for MSR would improve the performance.
Second, our proposed model classifies marine snow artifacts into two representative types.
Marine snow artifacts not fitted to these types are not well removed.
This limitation is described in the next section.

\subsection{Limitations}
Because our MSR benchmarking dataset is based on synthesizing marine snow artificially into real underwater images, some limitations exist.
First, the dataset is designed for removing marine snow artifacts and does not aim at achieving other underwater image enhancements such as color correction.
The design of a simultaneous restoration of MSR and color correction is an interesting area of future study.

\begin{figure}[h]\vspace{20pt}
\centering
    \subfigure{\includegraphics[width = 0.26\linewidth]{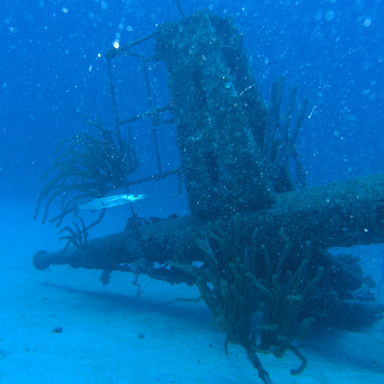}}
    \subfigure{\includegraphics[width = 0.26\linewidth]{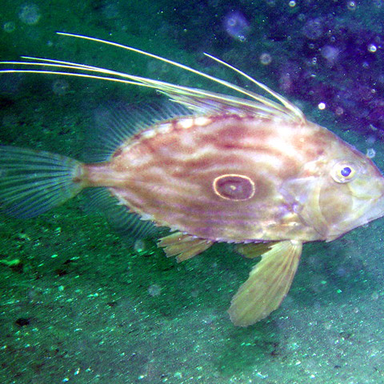}}
    \subfigure{\includegraphics[width = 0.26\linewidth]{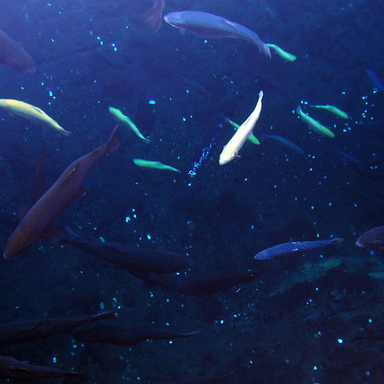}}\\
    \subfigure{\includegraphics[width = 0.26\linewidth]{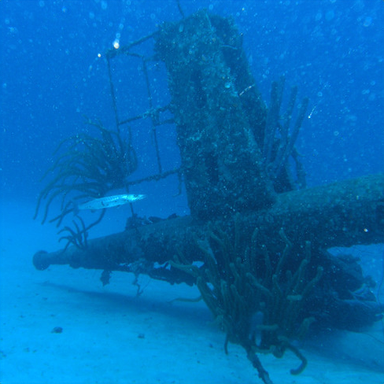}}
    \subfigure{\includegraphics[width = 0.26\linewidth]{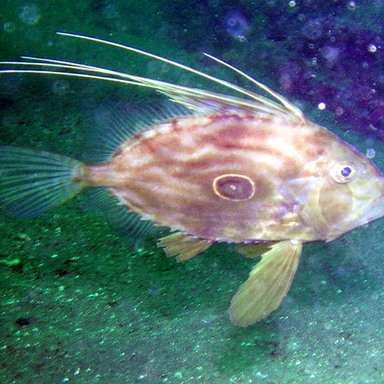}}
    \subfigure{\includegraphics[width = 0.26\linewidth]{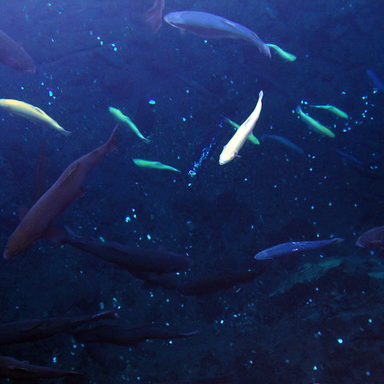}}
    \caption{Limitations. Top: Original images. Bottom: Restoration results by U-Net.}
  \label{unsuccessful_result}
\end{figure}

Although U-Net trained using the MSRB Dataset presents promising results, the dataset itself is imperfect.
Specifically, real MSR performances are occasionally limited if MSR artifacts are not fitted to our marine snow models.
Examples are shown in Fig. \ref{unsuccessful_result}, which compares real underwater images and their restoration results using U-Net.
Typically, U-Net fails to suppress artifacts when marine snow artifacts are larger and denser than our MSR benchmarking dataset.
Moreover, if the lighting conditions differ from the dataset, the current version of U-Net will fail.
Through a future study, this can be improved by carefully updating the MSRB Dataset to reflect various underwater scenes.


\section{Conclusions}
\label{sec:conclusion}
In this paper, the \emph{Marine Snow Removal Benchmarking Dataset} was proposed.
We mathematically modeled two representative marine snow artifacts and synthesized them in real underwater images.
Two MSR tasks, the removal of small-sized and various-sized marine snow artifacts were also proposed.
The first benchmarking results were also shown, which revealed the effectiveness of a deep neural network for MSR compared to median filter-based methods.
MSR for real underwater images indicates that our MSRB Dataset contributes to real MSR problems.
Our future studies will include improving the marine snow models as well as designing deep neural networks specifically designed for MSR.

{\small
\bibliographystyle{ieee_fullname}
\bibliography{sut_ref}
}

\end{document}